\documentclass[10pt,twocolumn,letterpaper]{article}

\usepackage{iccv}
\usepackage{times}
\usepackage{epsfig}
\usepackage{graphicx}
\usepackage{amsmath}
\usepackage{amssymb}
\usepackage{booktabs}
\usepackage{placeins}
\usepackage{wrapfig}
\usepackage[export]{adjustbox}
\usepackage{enumitem}
\usepackage{multirow}
\usepackage[accsupp]{axessibility} 
\usepackage{bbding}
\usepackage{pifont}
\usepackage{wasysym}
\iccvfinalcopy % *** Uncomment this line for the final submission

\def\iccvPaperID{3945} % *** Enter the ICCV Paper ID here

\ificcvfinal\fi

\begin{document}

%%%%%%%%% TITLE
%\title{Multi-class Counting via Exemplar-based Object Segmentation}
\title{Learning from Pseudo-labeled Segmentation for Multi-Class Object Counting}
%\title{Learning from Pseudo-labeled Segmentation \\for Multi-Class Class-Agnostic Object Counting}

\author{Jingyi Xu\\
Stony Brook University\\
{\tt\small jingyixu@cs.stonybrook.edu}
\and
Hieu Le\\
EPFL\\
{\tt\small minh.le@epfl.ch}
\and
Dimitris Samaras\\
Stony Brook University\\
{\tt\small samaras@cs.stonybrook.edu}
}

\maketitle
\ificcvfinal\thispagestyle{empty}\fi

%%%%%%%%% ABSTRACT
\begin{abstract}

Class-agnostic counting (CAC) has numerous potential applications across various domains. The goal is to count objects of an arbitrary 
category during testing, based on only a few annotated exemplars. 
In this paper, we point out that the task of counting objects of interest when there are multiple object classes in the image (namely, multi-class object counting) is particularly challenging for current object counting models. They often greedily count every object regardless of the exemplars. To address this issue, we propose localizing the area containing the objects of interest via an exemplar-based segmentation model before counting them. The key challenge here is the lack of segmentation supervision to train this model. To this end, we propose a method to obtain pseudo segmentation masks using only box exemplars and dot annotations. We show that the segmentation model trained on these pseudo-labeled masks can effectively localize objects of interest for an arbitrary multi-class image based on the exemplars. To evaluate the performance of different methods on multi-class counting, we introduce two new benchmarks, a synthetic multi-class dataset and a new test set of real images in which objects from multiple classes are present. Our proposed method shows a significant advantage over the previous CAC methods on these two benchmarks.

\end{abstract}

%%%%%%%%% BODY TEXT
\section{Introduction}

\begin{figure}[t]
\begin{center}
\includegraphics[width=0.82\linewidth]{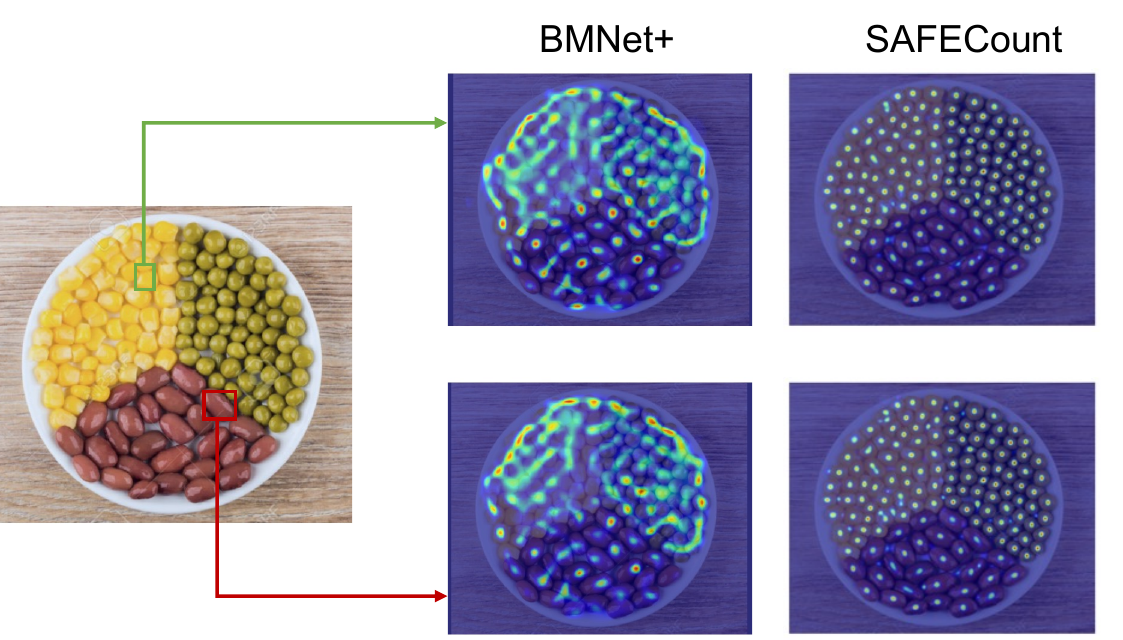}
%\makebox[0.45\linewidth]{(a) Few-shot Counting }
%\makebox[0.45\linewidth]{(b) Zero-Shot Counting}
\end{center} 
%\vspace{-7pt}
\caption{Visualizations of the density maps predicted by two recently proposed class-agnostic counting methods, \ie, BMNet+\cite{Shi2022SimiCounting} and SAFECount \cite{You2022FewshotOC}. They fail to count the objects of interest when multiple objects of different classes appear in the same image. %due to the training set containing mostly images with a single predominant category. 
%We propose a segment-then-count approach for counting class-agnostic objects in images in which multiple classes are present. We train a model to output an image segment that covers all objects similar to the input exemplar (no additional annotation required). 
} %
\label{fig:teaser} \vspace{-8pt}
\end{figure}

Class-agnostic counting (CAC) aims to infer the number of objects in an image, given a few object exemplars. Compared to conventional object counters that count objects from a specific category, \eg, human crowds \cite{Sam2022SSCrowd}, cars \cite{Mundhenk2016ALC}, animals \cite{Arteta2016CountingIT}, or cells \cite{Xie2018MicroscopyCC}, CAC can count objects of an arbitrary category of interest, which enables numerous applications across various domains.
%CAC has numerous potential applications across various domains as the counter can simply be used without additional data collection and training.
%Most of the existing methods focus on counting objects from specialized categories such as human crowds \cite{Sam2022SSCrowd}, cars \cite{Mundhenk2016ALC}, animals \cite{Arteta2016CountingIT}, and cells \cite{Xie2018MicroscopyCC}. These methods count only a single category at a time. 
%Recently, class-agnostic counting \cite{Ranjan2021LearningTC,Shi2022SimiCounting} has been proposed to count objects of arbitrary categories. Several human-annotated bounding boxes of objects are required to specify the objects of interest. 
%Although previous works on class-agnostic counting have shown the generalization ability to previously unseen classes,  we observe that they often fail to count the objects of interest when there are objects of multiple classes in the same image. 
% Explain that annotating multi-class training images is labor-expensive 
%Although tremendous progress has been made in class-agnostic counting lately,
%we observe that the existing counting models fail to count the objects of interest when objects of different classes appear at the same time (see Figure \ref{fig:teaser}). 

Most of the current CAC methods focus on capturing the intra-class similarity between image features \cite{Liu2022CounTRTG, Shi2022SimiCounting, Ranjan2021LearningTC, Gong2022ClassIntra}. For example, BMNet \cite{Shi2022SimiCounting} adopts a self-similarity module to enhance the feature's robustness against intra-class variations. Another recent approach, SAFECount \cite{You2022FewshotOC}, uses a similarity-aware feature enhancement framework to better capture the support-query relationship.
These methods perform quite well on the current benchmark, \ie FSC-147, in which images only contain objects from a single dominant class. However, we observe that when objects of multiple classes appear in the same image, these methods tend to greedily count every single object regardless of the exemplars (see Figure \ref{fig:teaser}). This issue greatly limits the potential applicability of these methods in real-world scenarios.
%, where images for counting often include multiple objects. 
A possible reason is that the current counting datasets only contain single-class training images, causing the counting models to overlook the inter-class discriminability due to the absence of multi-class training data.
%over-capture the intra-class similarity to minimize the counting error while .
%mostly focus on capturing the intra-class similarity between image patches \cite{Liu2022CounTRTG, Shi2022SimiCounting, Ranjan2021LearningTC, Gong2022ClassIntra}, but overlook the inter-class discriminability due to the absence of multi-class training data. %However, building such labeled multi-class datasets for counting is not an easy task, in terms of both collecting and annotating: it is non-trivial to collect images for counting with a large diversity of object categories, and annotating them can be even more costly since point annotation is required for instances from different classes.

%A natural solution to resolve this issue is to train the counting model with images containing objects of multiple classes. However, our experiments on a simple synthetic dataset with images containing objects from multiple categories show that .

%Note that this if there is an available real dataset ...

A natural solution to resolve this issue is to train the counting model with images containing objects of multiple classes. However, building such labeled multi-class datasets for counting is not an easy task: it is non-trivial to collect images for counting with a large diversity of categories, and annotating them is costly since point annotation is required for instances from different classes.
%Synthetic multi-class images become an alternative to supervising object counting models. 
An alternative way is to synthesize multi-class training data from single-class images.
By simply concatenating two or more images of different classes, we can easily create a large amount of multi-class data without additional annotation costs.
%
%Training on these synthetic data seems to be a feasible way to resolve the over-counting issue with reduced annotation cost.
%reduce annotation costs. 
However, our experiments show that %although this strategy can effectively reduce the counting error for multi-class images, 
while the model trained on these images indeed performs better on multi-class test images, 
the performance on single-class counting drops significantly (\ref{sec:comparison_synthetic}). %A possible explanation for this performance drop is that in order for the model 
This could be because, in order for the model to selectively count the objects of interest, it needs to recognize certain discriminative features that can distinguish between different classes. This will inevitably sacrifice some of its robustness against the variations within the same class. In other words, there is a trade-off between invariance and discriminative power for the counting model \cite{Varma2007LearningTD}. %In fact, our experiments (\ref{sec:trade-off}) show that after fine-tuning the model on our synthetic multi-class dataset, the inter-class separability of the extracted features is enhanced while the intra-class compactness is decreased. 

%Thus, instead of training one model to perform the task of single-class counting and multi-class counting simultaneously, our strategy is to compute an object mask to eliminate the non-target area first.
%and only count the objects within the target area. 
Due to this trade-off, instead of training an end-to-end model for multi-class counting, our strategy is to localize the area containing the objects of interest first and then count the objects inside. 
Given a multi-class image for counting and a few exemplars, our goal is to obtain a segmentation mask highlighting the regions of interest. Such an exemplar-based segmentation model can be easily trained if mask annotations are available. However, this is not the case for the current counting datasets \cite{Ranjan2021LearningTC, Hsieh2017DroneBasedOC}, and collecting such annotations is time-consuming and labor-intensive.
%However, such pixel-level segmentation masks are not available for our synthetic multi-class dataset, and collecting them is time-consuming and labor-intensive.
To this end, we propose a method to obtain pseudo segmentation masks using only box exemplars and dot annotations. We show that a segmentation model trained with only these pseudo-labeled masks can effectively localize objects of interest for multi-class counting.

%Ideally, the pseudo masks would cover all the objects that belong to the same class as the exemplars and not include any irrelevant object or background. 
We aim to obtain a mask covering all the objects that belong to the same class as the exemplars while not including any irrelevant object or background. 
%We use $K$-Means . 
We show that an unsupervised clustering method, $K$-Means, can be used for this purpose. In particular, given a synthetic multi-class image, a few annotated exemplars, and a pre-trained single-class counting model, we first represent each mask pixel with an image patch based on the receptive field of the network. Then we extract the feature embeddings for all the image patches as well as the provided exemplars and run $K$-Means clustering on them. %Those patches whose embeddings fall into the same cluster as the exemplar will be considered to contain the object of interest, 
%which form a mask indicating the locations of objects to count. which results in a positive label in the corresponding mask pixel, and vice versa. 
We consider the patches whose embeddings fall into the same cluster as the exemplar to contain the objects of interest and assign positive labels to the corresponding mask pixels. We assign negative labels otherwise.
Note that the output of $K$-Means is sensitive to the choice of $K$, which is hard to determine for each image.
In our case, we choose the $K$ that results in a pseudo mask that best benefits the pre-trained counting model, \ie, the counting model can produce the density map closest to the ground truth map after the pseudo mask is applied. 
%In our case, we choose the $K$ such that after applying the corresponding binary mask, the counting model can produce a density map closest to the ground truth density map. %In other words, we want to select the $K$ which leads to a segmentation mask that perfectly covers all the objects of interest and leads to the most accurate counting results.
The obtained pseudo masks can then be used as the supervision signal to train an exemplar-based segmentation model.

To evaluate the performance of different methods on multi-class counting, we introduce two new benchmarks, a synthetic multi-class dataset originating from FSC-147, and a new test set of real images in which objects from multiple classes are present. Our proposed method outperforms current counting methods by a large margin on these two benchmarks.

In short, our main contributions are:
\begin{itemize}%[leftmargin=*]
\setlength\itemsep{-.3em}
\item We identify a critical issue of the previous class-agnostic counting methods, \ie, greedily counting every object when objects of multiple classes appear in the same image, and propose a simple segment-and-count strategy to resolve it.
%\item We propose a clustering-based method to obtain pixel-level segmentation masks given only exemplars and train a segmentation model to approximate the optimal mask obtained from clustering.
%\item We propose a simple segment-and-count strategy for counting objects in images containing multiple counting object categories.
\item We propose a method to obtain pseudo-labeled segmentation masks using only annotated exemplars and use them to train a segmentation model.
%\item We verify the effectiveness of our method on both synthetic and real multi-class datasets, through extensive ablation studies and visualization results.% are provided to show the effectiveness of our method.
\item We introduce two benchmarks for multi-class counting, on which our proposed method outperforms the previous counting methods by a large margin.

\end{itemize}

\section{Related Work}

\begin{figure*}[!ht]
\begin{center}
\includegraphics[width=2\columnwidth]{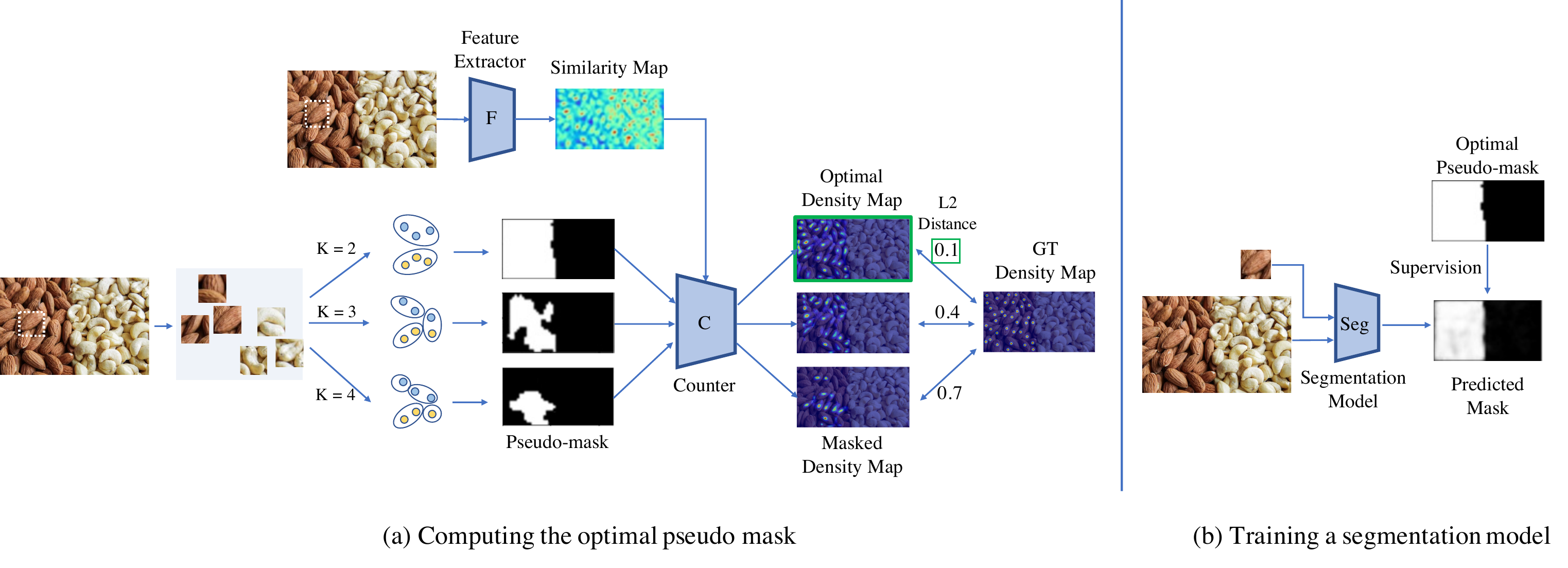} %\vspace{-1mm}
\caption{
Overview of our approach. We propose a method to obtain the pseudo segmentation masks using only box exemplars and dot annotations (a), and then use the obtained pseudo masks to train an exemplar-based segmentation model (b). Specifically, given a multi-class image and a few annotated exemplars, we crop a set of image patches, each of which corresponds to a mask pixel (we only visualize $6$ patches here for simplicity). 
%We extract the feature embeddings for all the cropped patches as well as the exemplars and run $K$-Means clustering on them. 
We run $K$-Means clustering on the feature embeddings extracted from all cropped patches and the exemplars.
Those pixels whose embeddings fall into the same cluster as the exemplar form an object mask indicating the image area containing the objects of interest. We find the optimal number of clusters, $K$, such that the counting model can produce the density map closest to the ground truth after the pseudo mask is applied. We use the obtained pseudo masks to train an exemplar-based segmentation model, which can then be used to infer the object mask given an arbitrary test image.
} 
\label{fig:overview}  
\end{center} \vspace{-7mm}

\end{figure*}

%\todo{read \cite{Yu2020EpisodeBasedPG}}

\label{sec:rw}
\subsection{Class-specific Object Counting} Class-specific object counting aims to count objects from pre-defined categories, such as humans \cite{Lian2019DensityMR,zhang2016singleCC,Zhang2019AttentionalNF,Wang2021NWPUCrowdAL,Sindagi2019PushingTF,Idrees2018CompositionLF,Abousamra2021LocalizationIT,Zhang2015CrosssceneCC,Sam2022SSCrowd,Zhang2022CaliFree,Xiong2022DiscreteConstrainedRF,Liu2022LeveragingSF,Wan2021AGL}, animals \cite{Arteta2016CountingIT}, cells \cite{Xie2018MicroscopyCC} and cars \cite{Mundhenk2016ALC,Hsieh2017DroneBasedOC}. Generally, there are two groups of class-specific counting methods: detection-based methods \cite{Chattopadhyay2017CountingEO,Hsieh2017DroneBasedOC,Laradji2018WhereAT} and regression-based methods \cite{Zhang2015CrosssceneCC,Cholakkal2019ObjectCA,Cholakkal2022TowardsPS, Wang2020DMCrowd,zhang2016singleCC,Chan2008PrivacyPC, Liu2019ContextAwareCC}. Detection-based methods apply an object detector on the image and count the number of objects based on the detected boxes. However, detection-based methods often struggle with detecting tiny objects.
Regression-based methods predict a density map for each input image, and the final result is obtained by summing up the pixel values. Both types of methods require a large amount of training data with rich training annotations. 
%Class-specific counters can perform well on trained categories.
Moreover, they can not be used to count objects of arbitrary categories at test time.

\subsection{Class-agnostic Object Counting} 
Class-agnostic object counting aims to count arbitrary categories given only a few exemplars \cite{Lu2018CAC,Ranjan2021LearningTC, Yang2021ClassagnosticFO,Shi2022SimiCounting,Gong2022ClassIntra,Nguyen2022fsoc,Liu2022CounTRTG,fsoc2023you,Arteta2014InteractiveOC}.
%GMN \cite{Lu2018CAC} uses a shared embedding module to extract feature maps for both query images and exemplars, which are then concatenated and fed into a matching module to regress the object count.  CFOCNet \cite{Yang2021ClassagnosticFO} convolves the feature maps from the exemplars over the query feature maps to obtain the density maps for object counting. 
Previous methods mostly focus on how to better capture the similarity between exemplars and image features.
For example, SAFECount \cite{You2022FewshotOC} uses a similarity-aware feature enhancement framework to better model the support-query relationship.
RCAC \cite{Gong2022ClassIntra} is proposed to enhance the counter's robustness against intra-class diversity.
%However, these methods tend to overlook the inter-class discriminability due to the lack of multi-class training data.
Nguyen \etal \cite{Nguyen2022fsoc} recently introduce new benchmarks for object counting, which contains images of objects from multiple classes, originating from the FSC-147 and LVIS \cite{Gupta2019LVISAD} datasets. However, these benchmarks are designed for the task of jointly detecting and counting object instances in complex scenes, where the central focus is on how to detect them accurately. 
%introduce a new task of few-shot object counting and detection, which outputs the object bounding boxes along with the total object count. 
%However, the new benchmarks are not suitable for evaluating the performance of density-based counting methods. This is because the objects in the new benchmarks are mostly in complex scenes, rather than densely distributed in the image. The key challenge, therefore, lies in how to detect them accurately.

%FamNet \cite{Ranjan2021LearningTC} adopts a similar way to do correlation matching and further applies test-time adaptation. Gong \textit{et al.} \cite{Gong2022ClassIntra} propose to use exemplar feature augmentation and edge matching to improve the counter's robustness against intra-class diversity.

\subsection{Unsupervised Semantic Segmentation} 
A closely related task to ours is unsupervised semantic segmentation \cite{Ji2019InvariantIC, Cho2021PiCIEUS,Ouali2020AutoregressiveUI, Chen2019UnsupervisedOS, Hamilton2022UnsupervisedSS, Hwang2019SegSortSB, VanGansbeke2021UnsupervisedSS, VanGansbeke2020SCANLT}, which aims to discover classes of objects within images without external supervision. 
IIC \cite{Ji2019InvariantIC} attempts to learn semantically meaningful features through transformation equivariance.
PiCIE \cite{Cho2021PiCIEUS} further improves on IIC's segmentation results by incorporating geometric consistency as an inductive bias.
Although these methods can semantically segment images without supervision, they typically require a large-scale dataset \cite{Caesar2016COCOStuffTA,Cordts2016TheCD} to learn an embedding space that is cluster-friendly. %Our method, on the other hand, does not require any representation learning step. Moreover, 
Moreover, the label space of semantic segmentation is limited to a set of pre-defined categories. In comparison, our goal is to localize the region of interest specified by a few exemplars, which can belong to an arbitrary class.

\section{Method}
\label{sec:method}

In order to perform multi-class object counting, our strategy is to compute a mask that can be applied to the similarity maps of a pre-trained counting model to localize the area containing the objects of interest and count the objects inside.
Figure \ref{fig:overview} summarizes our approach. We propose a method to obtain pseudo segmentation masks using only box exemplars and dot annotations, and then use these pseudo masks to train an exemplar-based segmentation model. Specifically, given a multi-class image and a few annotated exemplars, we tile the input image into different patches, each of which corresponds to a pixel on the mask. We run $K$-Means clustering on the feature embeddings extracted from all cropped patches and the exemplars. Those mask pixels whose corresponding patch embeddings fall into the same cluster as the exemplar will form an object mask indicating the image area containing the objects of interest. We find the optimal number of clusters, $K$, such that a pre-trained single-class counting model can produce the density map closest to the ground truth after the pseudo mask is applied. We use the obtained pseudo masks to train an exemplar-based segmentation model, which can then be used to infer the object mask given an arbitrary test image.
For the rest of the paper, we denote the pre-trained single-class counting model as the ``base counting model". Below we will first describe how we train this base counting model and then present the detail of our proposed multi-class counting method.
%Our method requires a pre-trained exemplar-based counting model. 

\subsection{Training The Base Counting Model} 
\label{sec:baseline_training}

We first train a base counting model using images from the single-class counting dataset \cite{Ranjan2021LearningTC}. 
Similar to previous works \cite{Ranjan2021LearningTC,Shi2022SimiCounting}, the base counting model uses the input image and the exemplars to obtain a density map for object counting. 
The model consists of a feature extractor $F$ and a counter $C$. Given a query image $I$ and an exemplar $B$ of an arbitrary class $c$, we input $I$ and $B$ to the feature extractor to obtain the corresponding output, denoted as $F(I)$ and $F(B)$ respectively. $F(I)$ is a feature map of size $d * h_I * w_I $ and $F(B)$ is a feature map of size $d * h_B * w_B $. We further perform global average pooling on $F(B)$ to form a feature vector $b$ of $d$ dimensions.

After this feature extraction step, we  obtain the similarity map $S$ by correlating the exemplar feature vector $b$ with the image feature map $F(I)$. 
Specifically, let $w_{(i,j)} = F_{(i,j)}(I)$ be the channel feature at spatial position $(i,j)$, $S$ can be computed by:
 \begin{equation}\label{eq:simi}
     S_{(i,j)}(I, B) = w_{(i,j)}^T b.
\end{equation}

In the case where $n$ exemplars are given, we  use Eq. \ref{eq:simi} to calculate $n$ similarity maps, and the final similarity map is the average of these $n$ similarity maps.

We then concatenate the image feature map $F(I)$ with the similarity map $S$, and input them into the counter $C$ to predict a density map $D$.
%The counter receives the channel-wise concatenation of the image feature map $F(X)$ and the similarity map $S$, and then predicts a density map $D_{pr}$. 
The final predicted count ${N}$ is obtained by summing over the predicted density map ${D}$:
\begin{equation} \label{eq:final_count}
 {N} = \sum_{i,j}D_{(i,j)}, \vspace{-2mm}   
\end{equation}
where ${D}_{(i,j)}$ denotes the density value for pixel $(i,j)$. 
%We adopt a conventional $l_2$ loss as the counting loss $L_\textnormal{count}$ to supervise the training of the counting model: 
The supervision signal for training the counting model is the $L_2$ loss between the predicted density map and the ground truth density map:

\begin{equation}\label{eq:counting_loss}
L_{\textnormal{count}} = \|D(I, B) - D^{*}(I,B)\|_2^2, 
\end{equation}
where $D^{*}$ denotes the ground truth density map.

\subsection{Multi-class Object Counting}
\label{sec:multi_class_counting}
\subsubsection{Pseudo-Labeling Segmentation Masks}
In this section, we describe our method to obtain pseudo-masks using only box exemplars and dot annotations. The mask is of the same size as the similarity map from the base counting model
%Since collecting a multi-label object counting dataset is a labor-intensive process,
%In order to perform multi-class object counting, we first create synthetic multi-class images from the existing single-class dataset. 
%Specifically, we randomly select two images belonging to different classes, crop a part from each image and then concatenate the two cropped parts horizontally. 
%The counting model trained with single-class images will fail on these synthetic images by counting every object regardless of the object exemplars.   
%We propose to resolve this issue by computing a coarse mask that can be applied on the similarity maps of the pre-trained counting model to exclude the non-target area. 
and each pixel on the mask is associated with a region in the original image. Ideally, the pixel value on the mask is 1 if the corresponding region contains the object of interest and 0 elsewhere. Specifically, 
%let $M(x, y)$ denotes the mask value at position $(x,y)$ and $r$ denotes the downsampling ratio, the corresponding region in the 
for the pixel from the mask $M$ at location $(i, j)$, we find its corresponding patch $p(i,j)$ in the input image centering around $(i_I, j_I)$, where $i_I = i*r + 0.5*r$ and $j_I = j*r + 0.5*r$. Here, $r$ is the downsampling ratio between the original image and the similarity map. The width and height of $p(i,j)$ are set to be the mean of the width and height of the exemplar boxes.

We denote $\mathbb{P} = \{p_1, p_2, ... p_n\}$ as a set of image patches, each of which corresponds to one pixel in the mask. The goal is to assign a binary label to each patch indicating if it contains the object of interest or not. To achieve this, we first extract the ImageNet features for all patches in $\mathbb{P}$ to get a set of embeddings $\mathbb{F} = \{f_1, f_2, ... f_n\}$. Then we compute the average of the embeddings extracted from the examplar boxes in this image, denoted as $f_B$. We run $K$-means on the union of $\{f_1, f_2, ... f_n\}$ and $\{f_B\}$. Those patches whose embeddings fall into the same cluster as $f_B$ will be considered to contain the object of interest, and result in a 1 value in the corresponding pixel of the mask. On the contrary, the pixel value will be 0 if the corresponding patch embedding falls into a different cluster as $f_B$.  Here $K$-Means groups similar objects together, which can serve our purpose of segmenting objects belonging to different classes.

It is worth noting that the number of clusters, denoted as $K$, has a large effect on the output binary mask and the final counting results. 
If $K$ is too small, too many patch embeddings will fall into the same cluster as the exemplar embedding and the counter will over-count the objects; if $K$ is set too high, too few embeddings will fall into the same cluster, which results in too many regions being masked out. 
%However, it is non-trivial to find the best $K$ given an arbitrary image. 
%without knowing how many classes exist in the image. 
In our case, we find the optimal $K$ for each image in the multi-class training set that results in the binary mask that minimizes the counting error. 
%Then we use all the obtained pseudo masks as the supervision signal to train a segmentation model.
Specifically, given a multi-class image $\bar{I}$ and an exemplar $\bar{B}$, let $S(\bar{I}, \bar{B})$ denote the similarity map outputted by the pre-trained counting model, and $M(\bar{I}, \bar{B})^k$ denote the mask obtained when the number of clusters is set to $k$. By applying $M(\bar{I}, \bar{B})^k$ on $S(\bar{I}, \bar{B})$, the similarity scores on the non-target area are set to a small constant value $\epsilon$ and the similarity scores on the target area remain the same:
 \begin{align} \label{eq:masking_map}
    S(\bar{I}, \bar{B})_{(i,j)}^k = \Bigg\{
        \begin{array}{ll}
       S(\bar{I}, \bar{B})_{(i,j)},  & \text{if } M(\bar{I}, \bar{B})_{(i,j)}^k = 1, \\ 
       \\
        %\text{min}(S(\bar{I}, \bar{B})),  & \text{otherwise.}
        \epsilon,  & \text{otherwise.}
        \end{array} \vspace{-2mm}
\end{align} 
We then input $S(\bar{I}, \bar{B})^k$ to the pre-trained counter $C$ to get the corresponding density map $D(\bar{I}, \bar{B})^k$. We find the optimal $k$ such that the $L_2$ loss between the predicted density map and the ground truth density map is the smallest:

\begin{equation}\label{eq:optimal_k}
k^* = \operatorname*{argmin}_k \|D(\bar{I}, \bar{B})^k - D^{*}(\bar{I})\|_2^2, 
\end{equation}
where $k^*$ denotes the optimal $k$ and $D^{*}(\bar{I})$ denotes the ground truth density map for input image $\bar{I}$. 

\subsubsection{Training Exemplar-based Segmentation Model}
%------------------------------------------------------------------------
After obtaining the optimal masks for all the images in the multi-class training set, we train a segmentation model $P$ to predict the pseudo segmentation masks based on the input image and the corresponding exemplar.  In particular, suppose we have a multi-class image $\bar{I}$ and an exemplar $\bar{B}$, we first input $\bar{I}$ and $\bar{B}$ to the segmentation model to get the corresponding feature map output $P(\bar{I})$ and $P(\bar{B})$. We then apply global average pooling on $P(\bar{B})$ to form a feature vector $v$. In the case where multiple exemplars are provided, we apply global average pooling on each $P(\bar{B})$ and the final vector $v$ is the average of all these pooling vectors. 

The predicted mask $M^p$ is obtained by computing the cosine similarity between $v$ and the channel feature at each spatial location of $P(\bar{I})$. Specifically, the value of the predicted mask at position $(i,j)$ is:
 \begin{equation}\label{eq:cos_simi}
     M^p_{(i,j)}(\bar{I}, \bar{B}) = \text{cos}(P_{(i,j)}(\bar{I})^T, v).
\end{equation}
The supervision signal for training this segmentation model is the $L_2$ loss between the predicted mask and the optimal mask obtained by finding the best $k$ with Eq. \ref{eq:optimal_k}:

\begin{equation}\label{eq:masking_loss}
L_{\textnormal{seg}} = \|M^p(\bar{I}, \bar{B}) - M^{*}(\bar{I}, \bar{B})\|_2^2, 
\end{equation}
where $M^{*}(\bar{I}, \bar{B})$ denotes the optimal mask under $k^*$.

\subsubsection{Inference on Multi-class Testing Data}
After the exemplar-based segmentation model is trained, we use it together with the pre-trained counting model to perform multi-class object counting. Given an input image for testing, we first input it to the feature extractor of the pre-trained counting model to get the corresponding similarity map. Then we use the segmentation model to predict a coarse mask where high values indicate the region of interets. We binarize this predicted mask with a simple threshold and apply it on the similarity map based on Eq. \ref{eq:masking_map}. The counter then take the masked similarity map as input and predict the density map and final object counts.
%Compared with running $K$-means during testing, using a segmentation model not only allows us to perform object counting on images with arbitrary number of classes, but also significantly reduces inference-time cost.

\section{Experiments}
\label{sec:exp}

\subsection{Implementation Details}

\textbf{Network architecture} 
For the {base counting model}, we use ResNet-50 as the backbone of the feature extractor, initialized with weights of a pre-trained ImageNet model. The backbone outputs feature maps of $1024$ channels. For each query image, the number of channels is reduced to $256$ using $1 \times 1$ convolution. For each exemplar, the feature maps are first processed with global average pooling and then linearly mapped to a $256$-d feature vector. The counter consists of $5$ convolution and bilinear upsampling layers to regress a density map of the same size as the query image. 
The segmentation model shares the same architecture as the backbone of the feature extractor. The output mask is of the same size as the similarity map from the base counting model.
 \def\subboxsize{0.45\textwidth}
 \begin{figure*}[ht!]
 \centering
\includegraphics[width=0.8\linewidth]{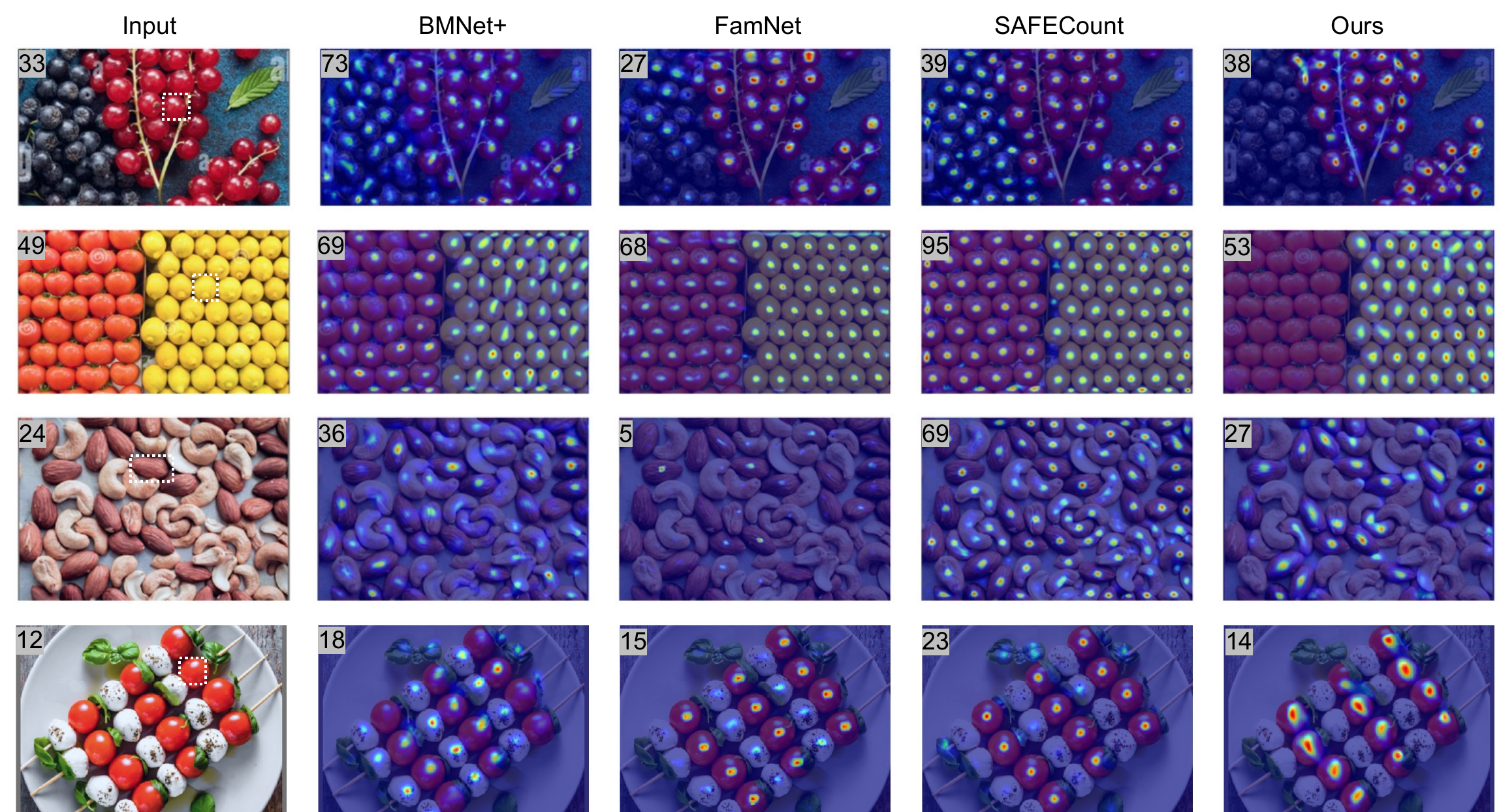}
 \caption{Qualitative results on our collected multi-class counting test dataset. We visualize a few input images, the corresponding annotated exemplar (bounded in a dashed white box) and the predicted density maps. Predicted object counts are shown at the top-left corner. Our predicted density maps can highlight the objects of interest specified by the annotated box, which will lead to more accurate object counts.
} 
\label{fig:img_visualization} \vspace{-2mm}
\end{figure*}

\begin{table*}[!ht] 
  \centering
\resizebox{0.6\textwidth}{!}{%
  \begin{tabular}{l|cccc|cccc}
    \toprule
   \multirow{2}{*}{Method} & \multicolumn{4}{c|}{Val Set} & \multicolumn{4}{c}{Test Set} \\
    & MAE & RMSE & NAE & SRE & MAE & RMSE & NAE & SRE \\
    \midrule
    %\multirow{2}{*}{GMN \cite{Lu2018CAC}} & GT & {29.66} & {89.81} & - & - & {26.52} & {124.57} & - & - \\
    %& RPN & {40.96} & {108.47} & - & - & {39.72} & {142.81} & - & - \\
    %\midrule
    %\multirow{2}{*} {FamNet} & GT & {24.32} & {70.94} & {22.56} & {101.54} \\
    % & RPN & {48.62} & {120.77} & {53.82} & {155.38}  \\
    %\midrule
   CounTR \cite{You2022FewshotOC} & 32.29 & 47.07 & 1.89 &  3.31  & 40.20 & 83.03  & 1.85 & 3.79 \\
   {FamNet \cite{Ranjan2021LearningTC}} 
     & {18.15} & {33.16} & 0.63 & 4.42 & {22.22} & {40.85}  & 0.79 & 9.29 \\
   {FamNet+ \cite{Ranjan2021LearningTC}} 
     & {27.74} & {39.78} & 1.33 & 7.29 & {29.90} & {43.59}  & 1.16 & 8.82 \\
     BMNet \cite{Shi2022SimiCounting} & 32.39 & 46.01 & 1.75 & 9.86 & 36.94 & 46.73 & 1.65 & 9.46 \\
     BMNet+ \cite{Shi2022SimiCounting} & 31.09 & 42.43 & 1.75 & 9.51 & 39.78 & 57.85 & 1.81 & 11.96 \\
     SAFECount \cite{You2022FewshotOC} & 22.58 & 34.68 & 1.21 &  \textbf{2.18}  & 26.44 & 40.68 & 1.14 & 2.89 \\
    %{FamNet} & {32.15} & {98.75} & {32.27} & {131.46} \\
    %{BMNet} & & {34.09} & {105.05} & {31.27} & {124.85} \\
    %{BMNet+} & & {35.15} & {106.07} & {34.52} & {132.64} \\
    %{BMNet+} & {23.01} & {72.61} & {22.43} & {128.12} \\
    Ours  
      %& 32.69 & 45.29 & 1.83 & 10.16 &  44.89 & 69.85 & 1.94 & 12.91  \\  
      & \textbf{14.34} & \textbf{26.03} & \textbf{0.61} & {4.48} & \textbf{11.13} & \textbf{16.96} & \textbf{0.41} & \textbf{2.80} \\
    \bottomrule
  \end{tabular}} \\ \vspace{1mm}
  \caption{ Quantitative comparisons on our synthetic multi-class dataset. Our proposed method outperforms the previous class-agnostic counting methods by a large margin, achieving the lowest mean average error on both validation and test set.
  }\label{tab:test_val}%
\end{table*}

\textbf{Dataset} We train the base counting model on the FSC-147 dataset. FSC-147 is the first large-scale dataset for class-agnostic counting. It includes $6135$ images from $147$ categories varying from animals, kitchen utensils, to vehicles. The categories in training, validation, and test sets have no overlap. We create synthetic multi-class images from FSC-147 dataset to train the segmentation model. Specifically, we randomly select two images belonging to different classes, crop a part from each image and then concatenate the two cropped parts horizontally. 
To evaluate the performance of multi-class counting on real images, 
%We test our model on synthetic multi-class images. Besides, 
we further collect a test set of $450$ multi-class images. 
For each image in this test set, there are at least two categories whose object instances appear multiple times. We provide dot annotations for $600$ groups of object instances.
The synthetic validation set and test set contain $1431$ and $1359$ images respectively. We test the trained model on both the synthetic multi-class images and our collected real multi-class images. 
 \def\subboxsize{0.44\textwidth}
 \begin{figure*}[ht!]
 \centering
\includegraphics[width=0.77\linewidth]{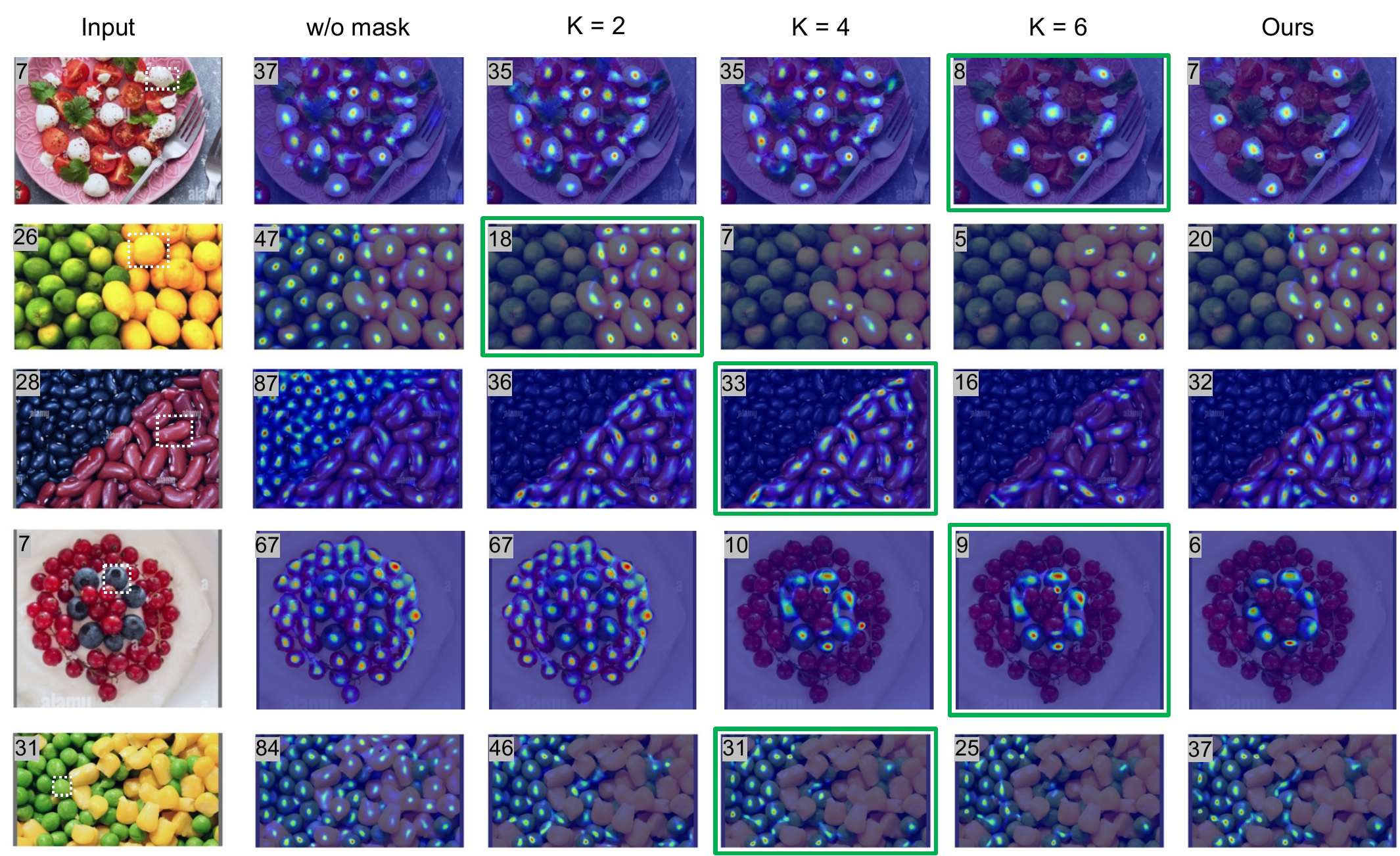}
 \caption{Qualitative analysis on the number of clusters. We visualize a few input images, the corresponding annotated exemplar (bounded in a dashed white box) and the density maps when using masks computed from $K$-means as well as predicted by our segmentation model. Predicted counting results are shown at the top-left corner. The density maps under the optimal $K$ are framed in \textcolor{green}{green}. The value of $K$ has a large effect on the counting results and the optimal $K$ varies from image to image.  
} 
\label{fig:k_visualization} \vspace{-2mm}
\end{figure*}

\textbf{Training details} 
Both the base counting model and the segmentation model are trained using the AdamW optimizer with a fixed learning rate of $10^{-5}$ and a batch size of $8$. The base counting model is trained for $300$ epochs and the segmentation model is trained for $20$ epochs. We resize the input query image to a fixed height of $384$, and the width is adjusted accordingly to preserve the aspect ratio of the original image.
Exemplars are resized to $128 \times 128$ before being fed into the feature extractor. We run $K$-means on the extracted patch embeddings to find the $K$ that leads to the optimal mask for each image. The embeddings are extracted from a pre-trained ImageNet backbone. The threshold for binarizing the segmentation mask is $0.6$ and the number of clusters $K$ ranges from $2$ to $6$. 
%he feature generation model is trained using the Adam optimizer and the learning rate is set to be $10^{-4}$. The semantic embeddings are extracted from CLIP \cite{Radford2021LearningTV}. For selecting the class-relevant patches, we randomly sample $450$ boxes of various sizes across the input query image and select $10$ patches whose embeddings are the $10$-nearest neighbors of the class prototype. The final selected patches are those that yield the top-$3$ smallest counting errors predicted by the error predictor.

\subsection{Evaluation Metrics}
%We use Mean Average Error (MAE) and Root Mean Squared Error (RMSE) to measure the performance of different object counters. Besides, we follow \cite{Nguyen2022fsoc} to report the Normalized Relative Error (NAE) and Squared Relative Error (SRE). 
For our collected multi-class test set, the counting error $\epsilon$ for image $i$ is defined as $\epsilon_i = |y_i-\hat{y_i}|$, where $y_i$ and $\hat{y_i}$ are the ground truth and the predicted number of objects respectively.
For our synthetic multi-class test set, the objects of interest are only present in the left / right-half part of the image. Ideally, the predicted number of objects should be close to the ground truth in the area of interest while being zero elsewhere. 
Thus, we define the counting error as $\epsilon_i = |y_i-\hat{y_i}| + \bar{\hat{y_i}}$, where $\hat{y_i}$ and $\bar{\hat{y_i}}$ denote the predicted number of objects in the interest area and non-interest area respectively. 

We use Mean Average Error (MAE), Root Mean
Squared Error (RMSE), Normalized Relative Error (NAE) and Squared Relative Error (SRE) to measure the performance of different object counters over all testing images. In particular, MAE = $\frac{1}{n} \sum_{i=1}^n \epsilon_i $; RMSE = $\sqrt{\frac{1}{n} \sum_{i=1}^n \epsilon_i ^2}$; NAE = $\frac{1}{n} \sum_{i=1}^n \frac{\epsilon_i}{y_i}$; SRE = $\sqrt{\frac{1}{n} \sum_{i=1}^n \frac{\epsilon_i^2}{y_i}}$ where $n$ is the number of testing images. %Compared with the absolute errors MAE and RMSE, the relative errors NAE and SRE better reflect the practical usage of visual counting \cite{Nguyen2022fsoc}.

\subsection{Comparing Methods}
We compare our method with recent class-agnostic counting methods, including CounTR (Counting TRansformer \cite{Liu2022CounTRTG}), FamNet (Few-shot adaptation and matching Network \cite{Ranjan2021LearningTC}), SAFECount (Similarity-Aware Feature Enhancement block for object Counting \cite{You2022FewshotOC}) and BMNet (Bilinear Matching Network \cite{Shi2022SimiCounting}). 

\begin{table}[!h] 
  \centering
\resizebox{0.34\textwidth}{!}{%
  \begin{tabular}{c|cccc}
    \toprule
    Method & MAE & RMSE & NAE & SRE \\
    \midrule
     CountTR \cite{You2022FewshotOC} & {24.73} & {45.16} & 1.62 & 3.10  \\
    FamNet \cite{Ranjan2021LearningTC} & {13.54} & {21.22} & 0.65 & 3.38  \\
    FamNet+ \cite{Ranjan2021LearningTC} & {19.42} & {38.46} & 0.95 & 6.13  \\
    BMNet \cite{Shi2022SimiCounting} & {21.92} & {37.09} & 1.18 & 1.68  \\
    BMNet+ \cite{Shi2022SimiCounting} & {25.55} & {40.35} & 1.36 & 1.81   \\
    SAFECount \cite{You2022FewshotOC} & {23.57} & {40.99} & 1.25 & 1.69  \\
    Ours & \textbf{6.97} & \textbf{13.03} & \textbf{0.37} & \textbf{0.54}  \\
    \bottomrule 
  \end{tabular}} \vspace{1mm}
  \caption{Quantitative comparisons on our collected multi-class dataset. Our proposed method has the lowest counting error compared with the previous class-agnostic counting methods.
  }\label{tab:real_testing} \vspace{-5mm}
\end{table}

\subsection{Results}
\textbf{Quantitative results.} 
Table \ref{tab:test_val} compares our proposed method with previous class-agnostic counting methods on our synthetic multi-class validation and test sets. (We include the results on single-class datasets in the Supp. Mat due to space limitations). The performance of all these methods drops significantly when tested on our synthetic dataset. The state-of-the-art single-class counting method CounTR \cite{Liu2022CounTRTG}, for example, shows a $20.34$ error increase \wrt validation MAE (from $13.13$ to $32.29$) and a $28.25$ error increase \wrt test MAE (from $11.95$ to $40.20$). Interestingly, we find that FamNet, which has the largest counting error on the single-class test set among these methods, performs best on our synthetic multi-class dataset. Unlike other methods, FamNet keeps the backbone of the counting model fixed without any adaptation, which prevents the model from over-capturing the intra-class similarity and greedily counting everything. This further validates that there is a trade-off between single-class and multi-class counting performance. Our proposed method outperforms the other methods by a large margin, achieving $14.34$ on validation MAE and $11.13$ on test MAE. 

Table \ref{tab:real_testing} shows the comparison with previous methods on our collected test set. Similarly, our proposed method significantly outperforms other methods by a large margin, as reflected by a reduction of $6.82$ \wrt MAE over FamNet and $14.77$ \wrt MAE over BMNet.

\textbf{Qualitative analysis.} 
In Figure \ref{fig:img_visualization}, we present a few input testing images, the corresponding annotated bounding box and the density maps produced by different counting methods. We can see that when there are objects of multiple classes present in the image, previous methods fail to distinguish them accurately, which often leads to over-counting. In comparison, the density map predicted by our method can highlight the objects of interest specified by the annotated box, even for the hard case where the objects are irregularly placed in the image (the 3rd row). 

\begin{table}[!h] 
  \centering
\resizebox{0.37\textwidth}{!}{%
  \begin{tabular}{c|cccccc}
    \toprule
    $K$ & 2 & 3 & 4 & 5 & 6 & Ours\\
    \midrule
    MAE & {15.13} & 10.77 & 8.17 & 7.98 & 8.03 & \textbf{6.97} \\
    RMSE & {28.09} & 20.71 & 15.38 & 14.93 & 15.31 & \textbf{13.03} \\
    NAE & {0.94} & 0.63 & 0.44 & 0.42 & 0.40 & \textbf{0.37} \\
    SRE  & {1.68} & 1.18 & 0.69 & 0.62 & \textbf{0.54} & \textbf{0.54} \\
    \bottomrule 
  \end{tabular}} \vspace{2mm}
  \caption{ Quantitative analysis on the number of clusters. Our proposed method outperforms $K$-Means under different values of $K$ on our collected multi-class test set.
  }\label{tab:ablation_k} \vspace{-2mm}
\end{table}

\definecolor{ashgrey}{rgb}{0.7, 0.75, 0.71}
\definecolor{aurometalsaurus}{rgb}{0.43, 0.5, 0.5}
\definecolor{babypink}{rgb}{0.96, 0.76, 0.76}
\definecolor{beaublue}{rgb}{0.74, 0.83, 0.9}
\definecolor{blue}{rgb}{0, 0, 1}
\definecolor{bluegray}{rgb}{0.4, 0.6, 0.8}
\newcommand{\mycolor}[1]{\color{black}{#1}}
\begin{table*}[!ht] 
  \centering
\resizebox{0.78\textwidth}{!}{%
  \begin{tabular}{l|c|cc|cccc}
    \toprule
   \multirow{2}{*}{Method} & Training & \multicolumn{2}{c|}{Multi Set} & \multicolumn{4}{c}{Single Set} \\
    & Set & Test MAE & Test RMSE & Val MAE & Val RMSE & Test MAE & Test RMSE   \\
    \midrule
    \multirow{2}{*}{BMNet+} & \mycolor{Single} & \mycolor{25.55} & \mycolor{40.35} & \mycolor{15.74} & \mycolor{58.53} & \mycolor{14.62} & \mycolor{91.83} \\
    & Single+Syn-multi & 11.44  & {23.22} & 24.24 & 73.42 & 20.89 & 99.04 \\
    \midrule
    \multirow{2}{*}{FamNet+ \cite{Ranjan2021LearningTC}} &  \mycolor{Single} &  \mycolor{19.42} &  \mycolor{39.78} &   \mycolor{23.75} &  \mycolor{69.07} &  \mycolor{22.08} &  \mycolor{99.54} \\
     & Single+Syn-multi & {11.31} & {18.84} &   29.45 & 94.33 & 26.93 & 116.12 \\
     \midrule
     \multirow{2}{*}{SAFECount \cite{Ranjan2021LearningTC}} &  \mycolor{Single} &  \mycolor{23.57} &  \mycolor{40.99}  &  \mycolor{14.42} &  \mycolor{51.72} &  \mycolor{13.56} &  \mycolor{91.30} \\
     & Single+Syn-multi & {9.80} & {32.40}  &  27.65 & 58.01 & 27.24 & 100.55 \\
     \midrule
     Ours & Single+Syn-multi & 6.97 & 13.03 & 18.55 & 61.12 & 20.68 & 109.14 \\
     %\multirow{2}{*}{BMNet \cite{Shi2022SimiCounting}} & GT & {18.29} & {124.02} & 0.26 & 4.39  \\
     %& RPN & {37.26} & {108.54} & {0.42} & {5.43}  \\
     %\midrule
    %\multirow{2}{*} {BMNet+ \cite{Shi2022SimiCounting}} & GT & {14.68} & {91.74} & {0.27} & {6.57}  \\
     %& RPN & {35.15} & {106.07} & {0.41} & {5.28}  \\
    \bottomrule
  \end{tabular}} \\ \vspace{1mm}
  \caption{ Comparison with training other class-agnostic counting methods (BMNet+, FamNet+ and SAFECount) using our synthetic multi-class images. Although the counting error on the multi-class test set is reduced, the performance on the single-class test set drops significantly for all three baseline methods. 
  }\label{tab:concat_training}  \vspace{-3mm}
\end{table*}

\section{Analyses}
\subsection{Comparison with Training with Synthetic Data}
\label{sec:comparison_synthetic}
Our strategy for multi-class counting is to compute a coarse mask to localize the image area of interest first and then count the objects inside with a single-class counting model. An alternative way is to train an end-to-end model for multi-class counting using images containing objects from multiple classes. In this section, we compare the performance of these two strategies. Specifically, we use our synthetic multi-class images to fine-tune three pre-trained single-class counting models: BMNet+ \cite{Shi2022SimiCounting}, FamNet+ \cite{Ranjan2021LearningTC} and SAFECount \cite{You2022FewshotOC}. Results are summarized in Table \ref{tab:concat_training}. As shown in the table, after fine-tuning on multi-class images, although the counting error on the multi-class test set is reduced, the performance on the single-class test set drops significantly for all three counting methods. Our method, in comparison, achieves the best performance for multi-class counting without sacrificing the performance on the single-class test set.

\subsection{Analysis on the Number of Clusters}
%In this section, we conduct experiments of using $K$-Means to generate segmentation masks at test time.
When running $K$-means, the number of clusters, $K$, has a large effect on the computed binary mask and the final counting results. However, it is non-trivial to determine $K$ given an arbitrary image. To resolve this issue, we first compute the optimal pseudo masks for the training images based on the dot annotations.
Then we train an exemplar-based segmentation model to predict the obtained pseudo masks. During testing, we can use the trained model to predict the segmentation mask based on exemplars.
%In our proposed method, we first find the optimal number of clusters which leads to the best binary mask, then we train a patch-based segmentation model to predict this binary mask based on the provided exemplar box. 
In this section, we provide analyses on how $K$ affects the final counting results and show a comparison with our proposed method.
\subsubsection{Quantitative Results}
We report the counting performance when computing masks by running $K$-means under different values of $K$ as well as using our predicted masks on the collected multi-class test set. Results are summarized in Table \ref{tab:ablation_k}. As $K$ goes from $2$ to $6$, both the MAE and RMSE decrease first and then increase, achieving the lowest when $K = 5$, \ie, $7.98$ \wrt MAE and $14.93$ \wrt RMSE. Using our predicted masks outperforms the performance under the best $K$ by $12.6\%$ \wrt MAE and $12.7\%$ \wrt RMSE, which demonstrates the advantages of using our trained segmentation model to predict the mask.

\subsubsection{Qualitative Results}

In Figure \ref{fig:k_visualization}, we visualize a few input images and the corresponding density maps when using masks computed from $K$-means as well as using masks predicted by our segmentation model. As can be seen from the figure, the choice of $K$ has a large effect on the counting results. If $K$ is too small, too many patch embeddings will fall into the same cluster as the exemplar embedding and the counter will over-count the objects (the 4th row when $K$ = 2); if $K$ is too large, too few embeddings will fall into the same cluster, which results in too many regions being masked out (the 2nd row when $K$ = 6). The optimal $K$ varies from image to image, and it is non-trivial to determine the optimal $K$ for an arbitrary image. Using our trained segmentation model, on the other hand, does not require any prior knowledge about the test image while producing more accurate masks and density maps based on the provided exemplars.

\subsection{Analysis on the Trade-off between Invariance and Discriminative Power}
\label{sec:trade-off}
We observe that there is a trade-off between single-class and multi-class counting performance. Our explanation is that when images contain objects from a single dominant class, the model will focus only on capturing the intra-class similarity while ignoring the inter-class discrepancy; when objects from multiple classes exist in the image, the model will focus more on the inter-class discrepancy in order to distinguish between them. 
To get a better understanding of this trade-off, we provide the detailed feature distribution statistics in Table \ref{tab:intra_inter}. 
Specifically, we measure the intra-class distance and inter-class distance of the exemplar features extracted from our baseline counting model before and after fine-tuning using the synthetic multi-class dataset. Intra-class distance refers to the mean of Euclidean distance between a feature embedding and the corresponding class's embedding center. Inter-class distance refers to the mean of the minimum distance between embedding centers. As shown in the table, after fine-tuning the model using the synthetic multi-class dataset, both intra-class distance and inter-class distance increase. Larger inter-class distance means features from different classes are more separable, suggesting a better discriminative power of the model;
larger intra-class distance means features within the same class are less compact, suggesting inferior robustness against within-class variations of the model. This trade-off between invariance and discriminative power makes it non-trivial to train one model to perform well on single-class counting and multi-class counting simultaneously.

\begin{table}[!ht] 
  \centering
\resizebox{0.45\textwidth}{!}{%
  \begin{tabular}{l|c|cccc}
    \toprule
     \multirow{2}{*}{Split} & Training &  \multirow{2}{*}{Intra} & \multirow{2}{*}{Inter} & Single & Syn-multi\\
        & Set &  &  & MAE & MAE   \\
    \midrule
    \multirow{2}{*}{Val} & Single & 2.35 & 1.12 & 18.55 & 32.46 \\
    & Single+Syn-multi & 2.90 & 1.30 & 32.36 & 25.74 \\
    \midrule
    \multirow{2}{*}{Test} & Single &  2.31 & 1.19 & 20.68 & 42.22 \\
     & Single+Syn-multi & 2.86 & 1.48 & 32.34 & 29.12 \\
    \bottomrule
  \end{tabular}} \\ \vspace{1mm}
  \caption{ Analysis on the trade-off between invariance and discriminative power of the counting model. After fine-tuning on our synthetic multi-class dataset, both the intra-class and inter-class distances of exemplar features become larger. 
  }\label{tab:intra_inter} \vspace{-2mm}
\end{table}

\section{Conclusion} 

In this paper, we identify a critical issue of the previous class-agnostic counting methods, \ie, greedily counting every object when objects of multiple classes appear in the same image. We show that simply training the counting model with multi-class data can alleviate this issue but often at the price of sacrificing the ability to count objects from a single class accurately.
%but will hurt the counting performance on the original single-class dataset. 
Thus, our strategy is to localize the area of interest first and then count the objects inside the area. To do this, we propose a method to obtain pseudo segmentation masks using only box exemplars and dot annotations. We show that a segmentation model trained with these pseudo-labeled masks can effectively localize the image area containing the objects of interest for an arbitrary multi-class image.
We further introduce two new benchmarks to evaluate the performance of different methods on multi-class counting, on which our method outperforms the other methods by a significant margin.

%\newpage
{\small
\bibliographystyle{ieee_fullname}
\bibliography{egbib}

}

\end{document}

%%%%%%%%% TITLE
%\title{Multi-class Counting via Exemplar-based Object Segmentation}
\title{Learning from Pseudo-labeled Segmentation for Multi-Class Object Counting\\ Supplementary Material}
%\title{Learning from Pseudo-labeled Segmentation \\for Multi-Class Class-Agnostic Object Counting}

\author{First Author\\
Institution1\\
Institution1 address\\
{\tt\small firstauthor@i1.org}
\and
Second Author\\
Institution2\\
First line of institution2 address\\
{\tt\small secondauthor@i2.org}
}

\maketitle
\ificcvfinal\thispagestyle{empty}\fi

%We discuss more later, these are just something on my mind now :-)
\section{Overview}
In this document, we provide additional experiments and analyses. In particular:
\begin{itemize}
    \item In Section \ref{sec:full_results}, we report the results of different counting methods on both the original single-class counting dataset and our synthetic multi-class dataset.
    \item In Section \ref{sec:ablation_pseudo_labeling}, we compare our proposed method with two alternative approaches to obtain pseudo masks.
    \item In Section \ref{sec:inference_time}, we report the time cost of running $K$-Means clustering and using our trained segmentation model.
     \item In Section \ref{sec:qual_number_clusters}, we provide additional qualitative analysis on the number of clusters.
      \item In Section \ref{sec:qual_results}, we provide additional qualitative comparisons of different counting methods on multi-class test images.
      \item In Section \ref{sec:details_test_set}, we provide more details about our collected multi-class test dataset.
\end{itemize}

\section{Results on Single-class and Multi-class Test Set} 
\label{sec:full_results}
In this section, we report the complete results of different class-agnostic counting methods on both the original FSC-147 dataset and our synthetic multi-class dataset. The images in our synthetic dataset are simply the concatenation of two random images from the single-class test set. Thus, a counter that counts objects based on the exemplars would perform similarly on these two test sets. However, as shown in Table \ref{tab:full_results}, the performance of all the previous methods drops significantly when tested on the synthetic multi-class test set. For example, BMNet+ \cite{Shi2022SimiCounting} shows a $15.35$ error increase \wrt validation MAE (from $15.74$ to $31.09$) and a $25.16$ error increase \wrt test MAE (from $14.62$ to $39.78$). Our proposed method outperforms
the other methods by a large margin, achieving 14.34 on validation MAE and 11.13 on test MAE.   

\begin{table*}[!ht] 
  \centering
\resizebox{0.67\textwidth}{!}{%
  \begin{tabular}{l|c|cccc}
    \toprule
   {Method} & Test Set & Val MAE & Val NAE & Test MAE & Test NAE \\
    \midrule
    \multirow{2}{*}{CounTR \cite{Liu2022CounTRTG}} & {Single} & {13.13} & {0.24} &  11.95 & 0.23 \\
    & Syn-multi & {32.29 (+ 19.16)} & {1.89 (+ 1.65)} & 40.20 (+ 28.25) & 1.85 (+ 1.62)\\
    \midrule
    \multirow{2}{*}{FamNet+ \cite{Ranjan2021LearningTC}} &  
    {Single} &  {23.75} & {0.52} & 22.08 & 0.44 \\
     & Syn-multi & {27.74 (+ 3.99)} & {1.33 (+ 0.81)} & 29.90 (+ 7.82) & 1.16 (+ 0.72) \\
     \midrule
         \multirow{2}{*} {BMNet+ \cite{Shi2022SimiCounting}} &  {Single} & {15.74} & {0.25} &  14.62 & 0.27 \\
     & Syn-multi & {31.09 (+ 15.35)} & {1.75 (+ 0.81)}  & 39.78 (+ 25.16) & 1.81 (+ 1.54)  \\
      \midrule
     \multirow{2}{*}{SAFECount \cite{You2022FewshotOC}} &  {Single} & {14.42} & {0.26} &  13.56 & 0.25 \\
     & Syn-multi & {22.58 (+ 8.16)} & {1.21 (+ 0.95)}  & 26.44 (+ 12.88) & 1.14 (+ 0.89)  \\
     %\multirow{2}{*}{BMNet \cite{Shi2022SimiCounting}} & GT & {18.29} & {124.02} & 0.26 & 4.39  \\
     %& RPN & {37.26} & {108.54} & {0.42} & {5.43}  \\
     \midrule
     \multirow{2}{*}{Ours} & {Single} & {18.55} & {0.30} &  20.68 & 0.36 \\
     & Syn-multi & {14.34 (- 4.21)} & {0.61 ( + 0.31)} & {11.13 (- 9.55)} & {0.41 ( + 0.05)}  \\
    \bottomrule
  \end{tabular}} \\ \vspace{1mm}
  \caption{ Results of different class-agnostic counting methods on the original FSC-147 dataset (denoted as ``Single") and on our synthetic multi-class counting dataset (denoted as ``Syn-multi"). 
  }\label{tab:full_results}  \vspace{-3mm}
\end{table*}
%\newpage
\section{Ablation on Pseudo-labeling Method}
\label{sec:ablation_pseudo_labeling}
In order to perform multi-class counting, we first generate pseudo-labeled segmentation masks from exemplars and dot annotations, and use the obtained masks to train a segmentation model. In this section, we compare our proposed method with two other strategies to obtain these pseudo segmentation masks.

\subsection{Comparing with Pseudo-labeling from Dot Annotations}
 An alternative way to obtain pseudo masks for training the segmentation model is to create pseudo boxes from dot annotations. In this subsection, we report the performance of this approach. Specifically, for each annotated dot from our synthetic multi-class dataset, we create a pseudo box centering around the dot. We experiment with three different sizes of pseudo box, \ie, the mean, minimum and maximum size of all exemplars, to generate the masks. The performance of the model trained on these masks is shown in Table \ref{tab:dot_annotations}. Our proposed method outperforms pseudo-labeling with dot annotations on both the synthetic multi-class test set and our collected real test set in all cases. For example, on our collected test set, the lowest MAE by pseudo-labeling from dot annotations is $8.57$, which is a $22.0\%$ error increase compared with our method. The results validate the advantage of our proposed method over pseudo-labeling using dot annotations.

\begin{table*}[!ht] 
  \centering
\resizebox{0.69\textwidth}{!}{%
  \begin{tabular}{c|c|cccc|cc}
    \toprule
    \multirow{2}{1cm}{Pseudo Masks} & \multirow{2}{1.3cm}{Box Size} & \multicolumn{4}{c|}{Syn-multi} &  \multicolumn{2}{c}{Real-multi}  \\
    &  & Val MAE & Val RMSE & Test MAE & Test RMSE & MAE & RMSE  \\
    \midrule
    w/o Mask & - & {32.46} & {45.25} & {42.22} & {59.95} & {24.68} & {41.70} \\
    \midrule
   \multirow{3}{*}{Dot Annos} & {Mean} & {18.93} & {35.30} & {12.48} & 21.51 & 9.26 & 19.23  \\
 & Min & {18.46} & 34.08 & 13.67 & 23.02 & 8.57 & 16.67 \\
  &  Max & 20.10 & 39.82 & 13.76 & 23.77 & 8.73 & 16.97  \\
  \midrule
  Ours & - & \textbf{14.34} & \textbf{26.03} & \textbf{11.13} & \textbf{16.96} & \textbf{6.97} & \textbf{13.03} \\
    \bottomrule
  \end{tabular}}  \vspace{2mm}
  \caption{ Comparison with pseudo-labeling from dot annotations. Our proposed method achieves lower counting errors on both the synthetic multi-class test set and our collected real test set in all cases. 
  }\label{tab:dot_annotations}  %\vspace{-3mm}
\end{table*}

\subsection{Comparing with Pseudo-labeling via Thresholding Similarity Map}

In this subsection, we compare our proposed method to obtain pseudo masks with pseudo-labeling via binarizing the similarity map between the image and the exemplar. Specifically, we input the image and the exemplar to a pre-trained ImageNet feature extractor to get the corresponding feature map. Then we do average pooling on the exemplar feature map to form a feature vector. We correlate the feature vector with the image feature map to obtain the similarity map, where a higher value indicates a higher similarity between the exemplar and the corresponding image area. We binarize this similarity map with a simple threshold to get the pseudo mask, which can then be used to localize the area of interest for multi-class counting. We experiment with different thresholds and the results are summarized in Table \ref{tab:simi_map}. We observe that the threshold for binarizing the similarity map plays a critical role in the final performance. When the threshold is set to $0.4$, the MAE and RMSE achieve the lowest on the synthetic test set, \ie, $20.91$ on Val MAE and $22.95$ on Test MAE. Our proposed method further improves the best results by a large margin, as reflected by a reduction of $6.57$ \wrt Val MAE and $11.82$ \wrt Test MAE.

\begin{table*}[!ht] 
  \centering
\resizebox{0.68\textwidth}{!}{%
  \begin{tabular}{c|c|cccc|cc}
    \toprule
    \multirow{2}{1cm}{Pseudo Masks} & \multirow{2}{1.5cm}{Threshold} & \multicolumn{4}{c|}{Syn-multi} &  \multicolumn{2}{c}{Real-multi}  \\
    &  & Val MAE & Val RMSE & Test MAE & Test RMSE & MAE & RMSE  \\
    \midrule
    w/o Mask & - & {32.46} & {45.25} & {42.22} & {59.95} & {24.68} & {41.70} \\
    \midrule
   \multirow{4}{*}{Similarity Map} & {0.2} & {31.35} & {41.90} & {38.63} & 53.00 & 24.94 & 37.60  \\
   & {0.4} & {20.91} & {34.18} & {22.95} & 32.74 & 11.08 & 19.78  \\
 & 0.6 & {27.12} & 44.88 & 27.52 & 40.09 & 17.93 & 29.76 \\
  &  0.8 & 30.50 & 47.79 & 32.60 & 44.07 & 20.67 & 31.85  \\
  \midrule
  Ours & - & \textbf{14.34} & \textbf{26.03} & \textbf{11.13} & \textbf{16.96} & \textbf{6.97} & \textbf{13.03} \\
    \bottomrule
  \end{tabular}}  \vspace{2mm}
  \caption{ Comparison with pseudo-labeling via binarizing the similarity map between the image and the exemplar. Our proposed method consistently outperforms binarizing the similarity map at different thresholds.  
  }\label{tab:simi_map} % \vspace{-3mm}
\end{table*}

\section{Inference Time Comparison}
\label{sec:inference_time}
%At inference time, using our trained segmentation model not only enables us to obtain more accurate masks, but also significantly reduces the time complexity compared with running $K$-Means clustering. 
Our trained segmentation model approximates the best outcome of $K$-Means, and is also much faster.
In Figure \ref{tab:inference}, we report the average time cost of running $K$-Means with different values of $K$ and using our trained segmentation model. As shown in the table, running $K$-Means results in a significantly higher time consumption compared with the baseline counting model. As the value of $K$ increases, the time consumption also increases accordingly. In comparison, using our trained segmentation model only results in marginal additional computation time, \ie, $0.015$s per real test image and $0.012$s per synthetic image.

\begin{table*}[!ht] 
  \centering
\resizebox{0.58\textwidth}{!}{%
  \begin{tabular}{c|c|ccccc|c}
    \toprule
   Test & \multirow{2}{1cm}{  w/o mask} & \multicolumn{5}{c|}{$K$-Means} & \multirow{2}{1cm}{  Ours}  \\
     Set &  & k = 2 & k = 3 & k = 4 & k = 5 & k = 6 &    \\
    \midrule
    {Real-multi} & {0.047} & {0.722} & {0.848} & 0.970 & 1.069
 & 1.161 & 0.061 \\
    Syn-multi & {0.021} & 0.767 & 0.862 & 0.924 & 1.012 & 1.061 & 0.033 \\
     %& RPN & {35.15} & {106.07} & {0.41} & {5.28}  \\
    \bottomrule
  \end{tabular}}  \vspace{2mm}
  \caption{The average time cost of running $K$-Means and using our segmentation model on the real test set and our synthetic test set. All results are in the unit of seconds. Our proposed method only takes around $30$ to $60$ ms, which is much faster than $K$-Means.
  %only results in marginal additional time cost while running $K$-Means clustering significantly increases the time complexity. 
  }\label{tab:inference}  %\vspace{-3mm}
\end{table*}

\section{Qualitative Analysis on the Number of Clusters}
\label{sec:qual_number_clusters}
\begin{figure*}[!ht]
\begin{center}
\hspace*{0.48cm}\includegraphics[width=0.94\columnwidth]{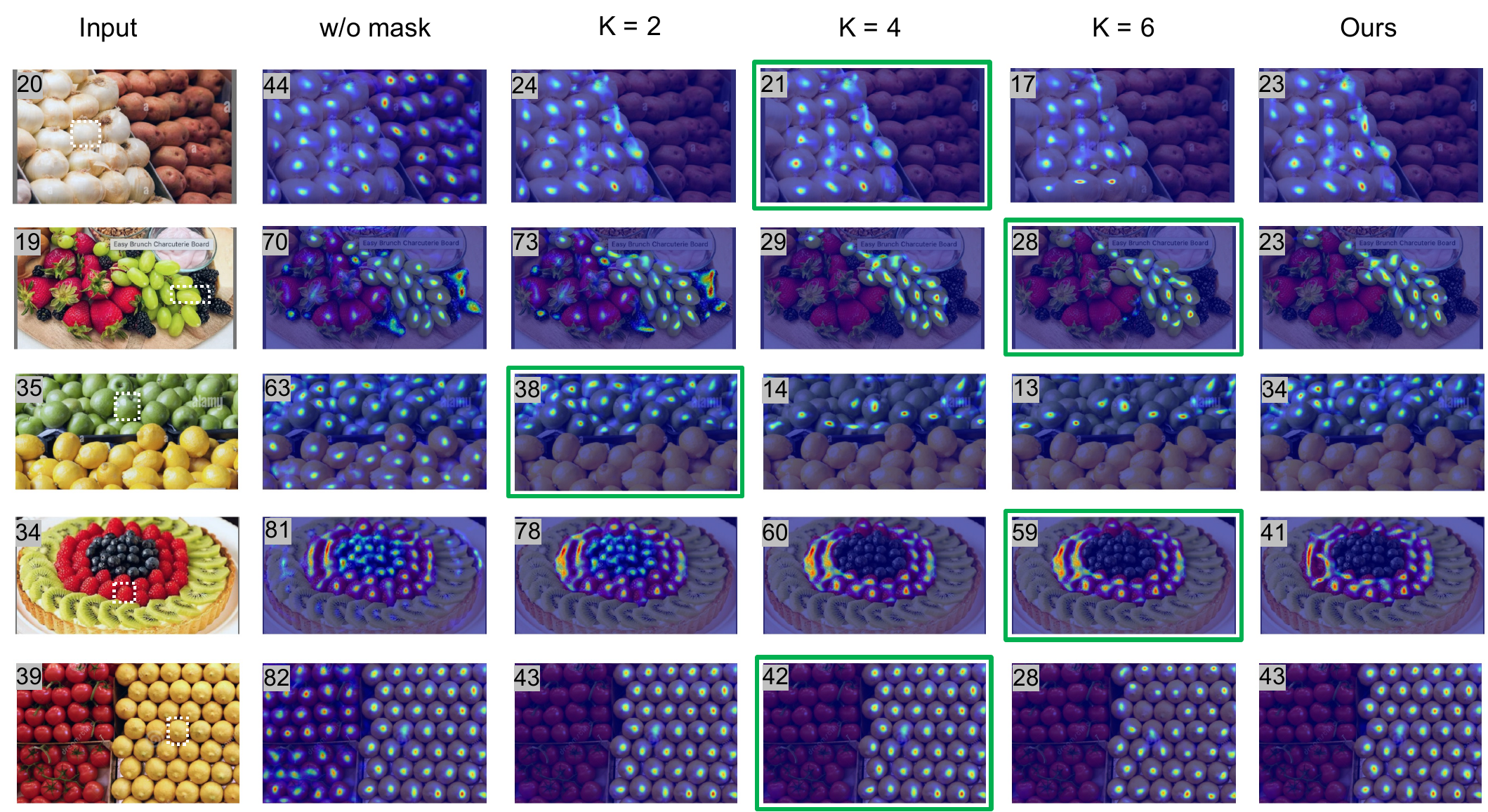}

\caption{Qualitative analysis on the number of clusters. We visualize a few input images together with the corresponding annotated exemplar (bounded in a dashed white box) and the density maps when using masks computed from $K$-means and masks predicted by our segmentation model. We only visualize one exemplar per image here for simplicity. Predicted counting results are shown in the top-left corner. The density maps under the optimal $K$ are framed in \textcolor{green}{green}. The value of $K$ has a large effect on the counting results and the optimal value of $K$ varies from image to image.  
} 
\label{fig:k_analysis} \vspace{-2mm}
\end{center}

\end{figure*}

In this section, we provide additional qualitative analysis on how the number of clusters, $K$, affects the final counting results. As shown in Figure \ref{fig:k_analysis}, we visualize a few input images and the corresponding density maps when using masks computed from $K$-means and using masks predicted by our segmentation model. The choice of $K$ has a large effect on the counting results. The optimal $K$ varies from image to image, and it is non-trivial to determine the optimal $K$ for an arbitrary image. Instead, using our trained segmentation model can consistently produce more accurate masks and density maps based on the provided exemplars.

\section{Qualitative Results}
\label{sec:qual_results}
\begin{figure*}[!ht]
\begin{center}
\hspace*{0.48cm}\includegraphics[width=0.94\columnwidth]{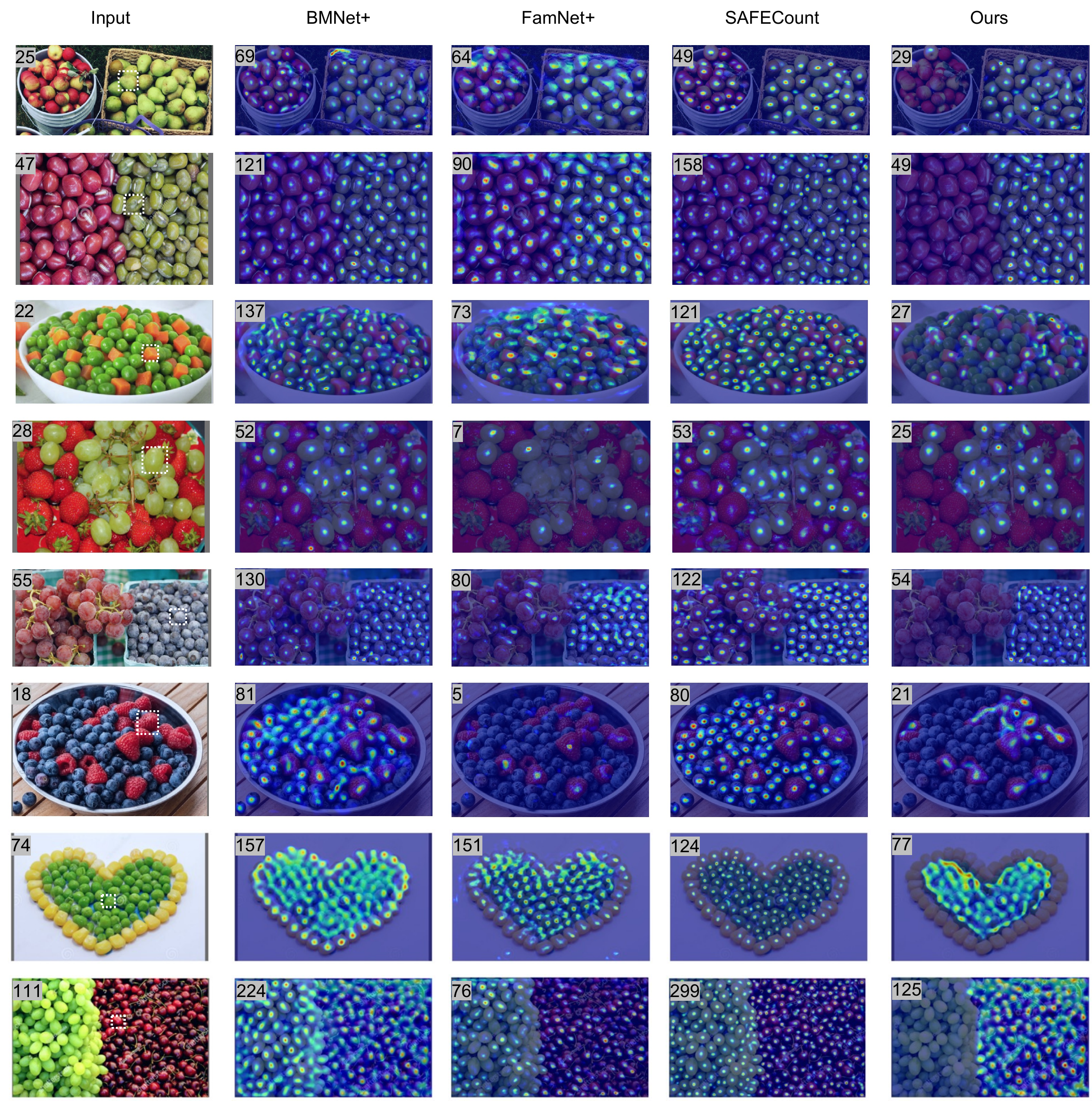}

\caption{ Qualitative results on our collected multi-class test set. We visualize a few input images, the corresponding annotated exemplar (bounded in a dashed white box) and the predicted density maps. Predicted object counts are shown in the top-left corner. Our predicted density maps can highlight the objects of interest specified by the annotated box, which will lead to more accurate object counts.
} 
\label{fig:multi}
\end{center}  \vspace{2mm}

\end{figure*}

In this section, we provide additional qualitative results of our proposed method in comparison with other class-agnostic counting methods. In Figure \ref{fig:multi}, we present a few input testing images, the corresponding annotated bounding box and the density maps produced by different counting methods. As can be seen, when there are objects of multiple classes present in the image, previous methods fail to distinguish them accurately, which often leads to over-counting. In comparison, the density map predicted by our method can highlight the objects of interest specified by the annotated box.

\section{Details on Real Multi-class Test Set}
\label{sec:details_test_set}
\begin{figure*}[!ht]
\begin{center}
\hspace*{0.48cm}\includegraphics[width=0.8\columnwidth]{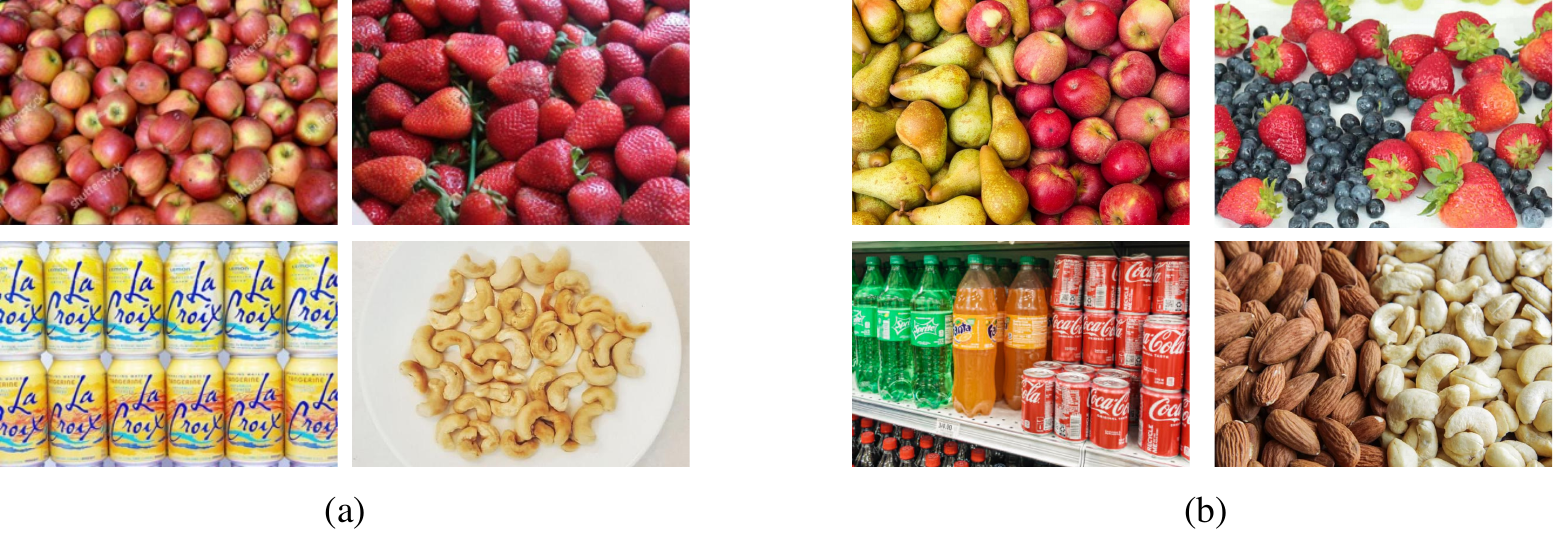} %\vspace{-2mm}
\caption{Sample images from FSC-147 dataset and our collected real test set. (a). Images from FSC-147 mostly contain objects from a single dominant class. (b). Images from our collected test set contain objects from multiple classes.
} 
\label{fig:dataset}
\end{center} %\vspace{-3mm}
\end{figure*}

Although the current dataset for class-agnostic counting, FSC-147 \cite{Ranjan2021LearningTC}, contains a large number of images with various object instances, the objects within each image are mostly from a single dominant class, as shown in Figure \ref{fig:dataset} (a). %Thus, it is not suitable for evaluating our proposed multi-class counting method.
However, in practice, there can be objects from multiple classes in the image, which is more challenging since the counter needs to selectively count only the objects of interest.
To evaluate the performance of different methods in this multi-class scenario, we collect and annotate a new test set of $450$ images, in which objects from different categories are present. For each image in this test set, there are at least two categories whose object instances appear multiple times. We provide dot annotations for $600$ groups of object instances. For each group, we randomly select $1$ to $3$ object instances as exemplar instances and annotate them with bounding boxes. Some sample images are shown in Figure \ref{fig:dataset} (b). We can see compared with the original FSC-147 dataset, the images in our collected dataset contain objects from multiple classes of different sizes and shapes. In some cases, objects from different classes are mixed together. These images are more likely to appear in real-world scenarios.

{\small
\bibliographystyle{ieee_fullname}
\bibliography{egbib}
}